\documentclass[runningheads]{llncs}
\usepackage{graphicx}
\usepackage{mdwtab}
\usepackage[flushleft]{threeparttable}
\usepackage{array}
\usepackage[caption=false,,font=footnotesize]{subfig}
\usepackage{amsmath, amssymb}
\usepackage[overload]{empheq}
\usepackage{algorithmic}
\usepackage[section]{placeins}

\begin{document}



\title{Joint data imputation and mechanistic modelling for simulating heart-brain interactions in incomplete datasets}

\titlerunning{Joint data imputation and mechanistic modelling}

\author{Jaume Banus\inst{1} \and Maxime Sermesant\inst{1} \and Oscar Camara\inst{2} \and Marco Lorenzi\inst{1}}

\authorrunning{J. Banus et al.}

\institute{Inria, Epione team, Université Côte d’Azur, Sophia Antipolis, France \email{jaume.banus-cobo@inria.fr}
\and PhySense, Department of Information and Communication Technologies, Universitat Pompeu Fabra, Barcelona, Spain}

\maketitle              
%


\begin{abstract}

The use of mechanistic models in clinical studies is limited by the lack of multi-modal patients data representing different anatomical and physiological processes. For example, neuroimaging datasets do not provide a sufficient representation of heart features for the modeling of cardiovascular factors in brain disorders. To tackle this problem we introduce a probabilistic framework for joint cardiac data imputation and personalisation of cardiovascular mechanistic models, with application to brain studies with incomplete heart data. Our approach is based on a variational framework for the joint inference of an imputation model of cardiac information from the available features, along with a Gaussian Process emulator that can faithfully reproduce personalised cardiovascular dynamics. Experimental results on UK Biobank show that our model allows accurate imputation of missing cardiac features in datasets containing minimal heart information, e.g. systolic and diastolic blood pressures only, while jointly estimating the emulated parameters of the lumped model. This allows a novel exploration of the heart-brain joint relationship through simulation of realistic cardiac dynamics corresponding to different conditions of brain anatomy. 

\end{abstract}
\section{Introduction}

Heart and brain are characterized by several common physiological and pathophysiological mechanisms \cite{Doehner2018}. The study of this multi-organ relationship is of great interest, in particular to better understand neurological diseases such as vascular dementia or Alzheimer's disease. The development of computational models simulating heart and brain dynamics is currently limited by the lack of databases containing information for both organs. Neuroimaging datasets often provide a limited number of cardiac-related measurements, usually restricted to brachial diastolic and systolic blood pressure (DBP, SBP) \cite{epstein2013cognitive}. These quantities provide a limited assessment of the cardiac function, thus compromising the possibility of further analysis of cardiovascular factors in brain disorders. 

On the contrary, the availability of a rich set of cardiac information in heart studies allows the use of cardiovascular models to estimate descriptors of the cardiac function that are not possible to measure in-vivo, such as contractility or stiffness of the heart fibers. These models optimize the parameters through data-assimilation procedures to reproduce the observed clinical measurements \cite{multifidelity}, but usually do not include neurological factors. The ability to jointly account for cardiovascular descriptors and brain information is key to gather novel insights about the relationship between heart dynamics and brain conditions. 

Most of current studies relating heart and brain are based on statistical association models, such as based on multivariate regression \cite{Cox2019}. While this kind of analysis allows to easily formulate and test association hypotheses, it usually  offers a limited interpretation of the complex relationship between organs. This issue is generally addressed by mechanistic modeling, allowing deeper insights on physiological and biomechanical aspects. These models allow for example to describe the brain vasculature, and to quantify physiological aspects such as blood flow auto-regulation effects in the brain \cite{Acosta2018}, up to the simulation of the whole-body circulation with detailed compartmental components \cite{Blanco2015}. Although these approaches offer a high level of interpretability, they are usually severely ill-posed and  require large data samples and arrays of measurements to opportunely tune their parameters. 

To bridge the gap between data-driven and mechanistic approaches to heart-brain analysis, in this work we propose 
to learn cardiovascular dynamics from brain imaging and clinical data  by leveraging on large-scale datasets with missing cardiac information. This is achieved through an inference framework composed of two nested models accounting respectively for the imputation of missing cardiac information conditioned on the available cardiac and brain features, and for a Gaussian Process emulator that mimics the behavior of a lumped cardiovascular model. This setting allow us to formulate a probabilistic end-to-end generative model enabling imputation of missing measurements and estimation of cardiovascular parameters given a subset of observed heart and brain features. 

Results on real data from the UK Biobank show that our framework can be used to reliably estimate and simulate 
cardiac function from datasets in which we have minimal cardiovascular information, such as SBP and DBP only. Moreover, the proposed framework allows novel exploration of the joint heart-brain relationship through the simulation of realistic cardiac dynamics corresponding to different scenarios of brain anatomy and damage.

\section{Methods}

\subsection{Problem statement}

\begin{figure}[!hbt]
\centering
\includegraphics[width=0.3\textwidth]{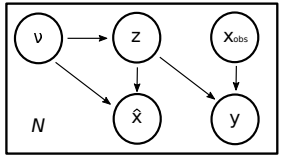}
\caption{Graphical model of our framework. From a latent variable $z$ we generate the unobserved features $\hat{x}$ conditioned on the variables $\nu$ and we estimate $y$ via Gaussian process regression. During inference $x_{obs}$ and $\nu$ are used to estimate the approximated posterior $q_{\phi}(z|x_{obs}, \nu)$.} \label{fig:graphical_model}
\end{figure}

We denote by $\nu$ the vector representing brain image-derived phenotypes (IDPs) and clinical information such as age or body surface area (BSA), and by $x$ the vector of cardiac IDPs and blood pressure measurements. The vector $x$ can be represented as $x=\{\hat{x}, x_{obs}\}$ where $\hat{x}$ represents the unobserved information we wish to impute and $x_{obs}$ the observed one. Moreover, we assume that for each observation $x$ a corresponding set of parameters $y$ of the associated lumped model is available. 
We would like to learn a generative model in which we assume that the unobserved measurements are generated by a latent random variable $z$ conditioned on the variables $\nu$. Hence, our generative process can be seen as sampling from a distribution $p(z|\nu)$ and then obtaining $\hat{x}$ with probability $p(\hat{x}|z, \nu)$. Due to the association between cardiac IDPs, $x$, and lumped model parameters $y$, we also assume that the latter are dependent from $\nu$ through $z$. Our graphical model is shown in Figure \ref{fig:graphical_model}, while the evidence lower bound (ELBO) of the joint data marginal $p(y, \hat{x} | x_{obs}, \nu)$ writes as:

\begin{equation} \label{eq:ELBO_JOINT}
    \begin{aligned}
        \log p(y, \hat{x}|x_{obs},\nu) & = \int_{}{} \log p(y,\hat{x}|x_{obs}, z, \nu)p(z|\nu)dz\\ 
        &\geq \mathbb{E}_{q_{\phi}(z|x_{obs}, \nu)} \log p(y,\hat{x}|x_{obs}, z, \nu)\\
        &-KL(q_{\phi}(z|x_{obs}, \nu)||p_{\theta}(z | \nu))\\
        &=\mathbb{E}_{q_{\phi}(z|x_{obs}, \nu)} \log p_{\omega}(y|x_{obs},z) \\
        &+ \mathbb{E}_{q_{\phi}(z|x_{obs}, \nu)} \log p_{\theta}(\hat{x}|\nu, z)\\ 
        &- KL(q_{\phi}(z|x_{obs}, \nu)||p_{\theta}(z | \nu)) \equiv \mathcal{L}(\theta, \phi, \omega; x_{obs}, \nu)
    \end{aligned}
\end{equation}

The approximation of the posterior distribution by $q_{\phi}(z|x_{obs}, \nu)$ defines the optimization of the ELBO through variational inference. The variational distributions $q_{\phi}(z|x_{obs}, \nu)$ and $p_{\theta}(\hat{x}|\nu, z)$ are parametrized respectively with parameters $\phi$ and $\theta$. The term $\log p_{\omega}(y|x_{obs},z)$ in the ELBO denotes the emulator which approximates the mechanistic behavior of the lumped model via Gaussian Process regression parametrized by $\omega$. The choice of a GP as emulator is motivated by the uncertainty of the data, hence it is desirable to obtain a distribution of interpolating functions rather than a single deterministic function. Moreover, GPs have already been proved to be valid emulators of 1D mechanistic vascular models \cite{Melis2017}. The GP allow us to sample functions $f(x)$ from a given prior parameterized by a mean $\mu(x)$ and covariance $\Sigma(x)$, i.e: $f(x) \sim \mathcal{N}(\mu(x), \Sigma(x))$ to obtain the marginal $y \sim \mathcal{N}(\mu(x), \Sigma(x) + \sigma^{2}I)$. The prior mean $\mu(x)$ is here set to 0 and we use a radial-basis function (RBF) kernel for the covariance:

\begin{equation} \label{eq: GP_Kernel}
    k^{j}(x_{i}, x'_{i}) = \alpha^{2}_{j}exp\left(-\frac{(x_{i}-x_{i}')^{2}}{2\beta^{2}_{i}}\right),
\end{equation}

Since our data is multi-dimensional, $k^{j}(x_{i},x'_{i})$ is the kernel for the $j^{th}$ target and the $i^{th}$ predictor. The hyper-parameters of the kernel $\omega = \{\alpha, \beta\}$ represent the output amplitude $\alpha$ and length scale $\beta$ of the sampled functions. The goal during training is to learn the hyper-parameters that maximize the marginal likelihood of the observed data $y$ . 

The second term of the ELBO is related to the imputation of $\hat{x}$. The term $\log p_{\theta}(\hat{x}|\nu, z)$ denotes the log-likelihood of the imputed features, which can be seen as the reconstruction error. The last term $KL(q_{\phi}(z|x_{obs}, \nu)||p_{\theta}(z | \nu))$ is the Kullback–Leibler divergence between variational approximation and prior for z, that can be expressed in a closed form given that both distributions are Gaussians. The imputation scheme is equivalent to a conditional variational autoencoder (CVAE) which has become a popular approach to feature imputation \cite{Ivanov2019}. Fast and efficient optimization in our model is possible by means of stochastic gradient descent thanks to the closed form for data fit and KL terms, the use of the reparametrisation trick and Monte Carlo sampling. 

\subsection{Data processing and cardiovascular model}

From UK Biobank we selected a subset of 3445 subjects for which T1, T2 FLAIR magnetic resonance images (MRI) and several brain and cardiac IDPs were available. Among the brain IDPs we used total grey matter (GM), total white matter (WM) and ventricles volumes. We used T1 and T2 FLAIR images to obtain the number of white matter hyper-intensities (WMHs) and their total volume relying on the lesion prediction algorithm (LPA), available from the lesion segmentation toolbox (LST)~\cite{LPA_algorithm} of SPM \footnote{https://www.fil.ion.ucl.ac.uk/spm}. WMHs are a common indicator of brain damage of presumably vascular origin \cite{Wardlaw2015}. We combined WM and GM volumes into a single measurement that we denoted as brain volume. The WMHs total volume and number of lesions presented a skewed distribution, and were Box-Cox transformed prior to the analysis. Regarding the cardiac IDPs we selected stroke volume (SV), ejection fraction (EF) and end-diastolic volume (EDV) for the left ventricle. All brain-related volumes were normalized by head size. Besides IDPs, we had access to blood pressure measurements (DBP, SBP) and socio-demographic features such as age and body surface area (BSA). We used DBP and SBP to compute MBP as $MBP=DP+(SP-DP)/3$. Next, we used the lumped cardiovascular model derived in \cite{Caruel2014} to obtain additional indicators of the cardiac function. In particular we estimated the contractility of the main systemic arteries ($\tau$), peripheral resistance ($R_{p}$) , the radius of the left ventricle  ($R_{0}$), contractility of the cardiac fibers ($\sigma_{0}$) and their stiffness ($C_{1}$). The parameters of the model were selected based on the available clinical data. To obtain the target values for the emulator, the data-assimilation procedure was carried out according to the approach presented in \cite{Banus2019,Mollero2018}. 

\subsection{Experiments}

The data was split in two sets: one containing the full-information (2309 subjects), and one in which cardiac IDPs ($\hat{x}$) and the estimated model parameters ($y$) were removed (1136 subjects,  Table 1). The quality of imputation was compared to conventional methods such as mean, median and k-nearest neighbors (KNN) imputation. We subsequently assessed the relationship learnt by our model between brain features and cardiovascular parameters through simulation. Starting from the mean values of each parameter we sampled along the dimension of the different conditional variables $\nu$ that we used to parameterize the prior. This procedure allowed us to assess their influence in the inferred simulation parameters. 

\begin{table}[!h]
\setlength\tabcolsep{5pt} 
\caption{Variables used in our framework. Mean blood pressure (MBP), diastolic blood pressure (DBP), stroke volume (SV), end-diastolic volume (EDV), ejection fraction (EF), heart fibers contractility ($\sigma_{0}$), ventricle size ($R_{0}$), heart fibers stiffness ($c_{1}$), peripheral resistance ($R_{p}$) and aortic compliance ($\tau$)}
\label{tab:population_statistics}
\centering
\begin{tabular}[C]{ l l l l} \hlx{hv}
Input ($x_{obs}$) & Condition ($\nu$) & Imputed ($\hat{x}$)  & Predicted ($y$) \\ \hlx{vhhv}
\multicolumn{1}{l|}{MBP} & \multicolumn{1}{l|}{Brain volume} & \multicolumn{1}{l|}{SV}  & \multicolumn{1}{l}{$\sigma_{0}$} \\
\multicolumn{1}{l|}{DBP} & \multicolumn{1}{l|}{Ventricles volume} & \multicolumn{1}{l|}{EDV}  & \multicolumn{1}{l}{$R_{0}$} \\
\multicolumn{1}{l|}{} & \multicolumn{1}{l|}{WMHs volume} & \multicolumn{1}{l|}{EF}  & \multicolumn{1}{l}{$c_{1}$} \\
\multicolumn{1}{l|}{} & \multicolumn{1}{l|}{Num. WMHs} & \multicolumn{1}{l|}{}  & \multicolumn{1}{l}{$R_{p}$} \\
\multicolumn{1}{l|}{} & \multicolumn{1}{l|}{BSA} & \multicolumn{1}{l|}{}  & \multicolumn{1}{l}{$\tau$} \\
\multicolumn{1}{l|}{} & \multicolumn{1}{l|}{Age} & \multicolumn{1}{l|}{}  & \multicolumn{1}{l}{} \\
\hlx{vh}
\end{tabular}
\end{table}

\section{Results}

\subsubsection{Data imputation and regression.}

\begin{figure}[!hbt]
\centering
\includegraphics[width=0.95\textwidth]{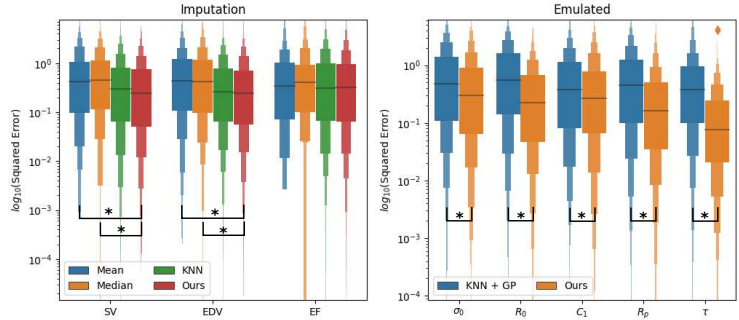}
\caption{Mean squared error (MSE) a) of the imputation of missing cardiac measurements ($\hat{x}$) b) of the estimated parameters of the emulated cardiovascular lumped model ($y$). * denotes that the MSE distributions are significantly different with respect to our method according to the Wilcoxon rank-sum test using a significance level of $a=0.05$, Bonferroni corrected by multiple comparison.} \label{fig:errors}
\end{figure}

We assessed the performance of our model by measuring the mean squared error (MSE) on the testing data, for both the imputation of missing cardiac information and the emulation of the lumped model parameters. The results in Figure \ref{fig:errors}a show that our method gives significantly better estimates than mean and median imputation, and comparable predictions to KNN, for which the optimal number of neighbors was optimized through 10-fold cross-validation and corresponds to $K=10$. At the same time the emulator consistently gives low errors for the parameters' estimation (Figure \ref{fig:errors}b). In supp. Figure 3 we present the most relevant predictors for each emulated feature based on their $\beta$ values, while a qualitative comparison between the distributions of imputed and emulated features compared to the ground truth data is available in in supp. Figure 4.

\subsubsection{Cardiovascular dynamics simulation and model plausibility.}

\begin{figure}[!hbt]
\centering
\includegraphics[width=0.88\textwidth]{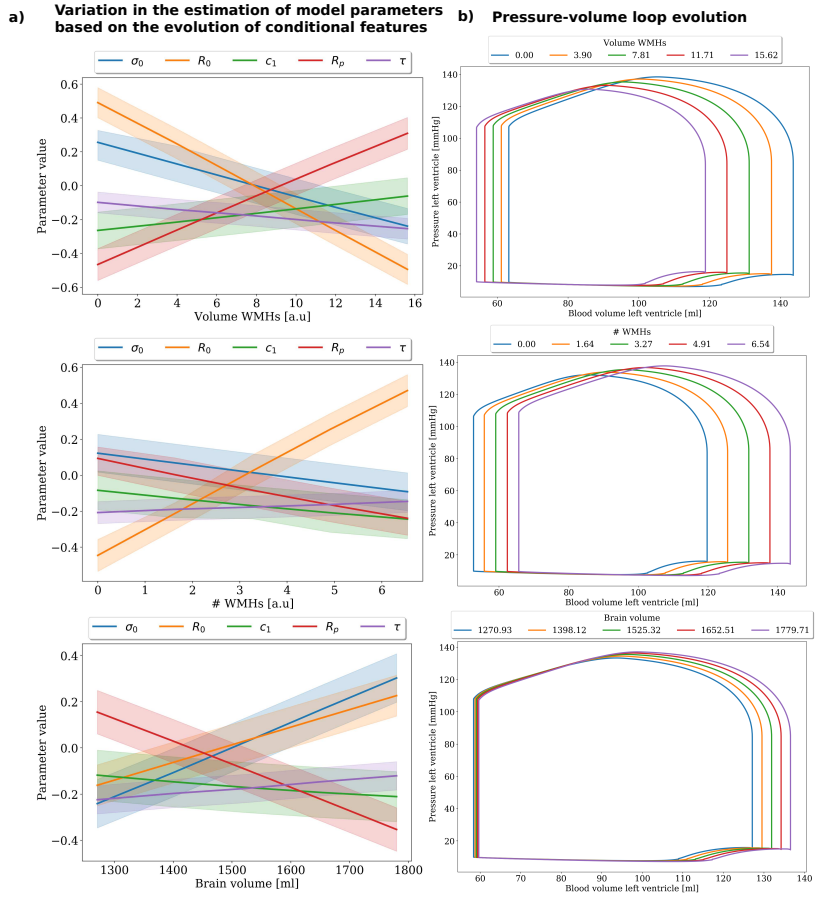}
\caption{a) Inferred model parameters and their respective confidence interval as we sample along the dimension of the different conditional variables $\nu$ while keeping the other elements of the generative framework constant. b) Pressure-volume loops generated by the the cardiovascular lumped model given the mean inferred parameters.} \label{fig:confidence_intervals}
\end{figure}

In Figure \ref{fig:confidence_intervals}a we observe the change in the predicted parameters of the cardiovascular model as we sample along the range of values of the different conditional variables $\nu$. Figure \ref{fig:confidence_intervals}b shows the pressure-volume (PV) loops generated by the lumped model using the inferred parameters. The simulated PV loops highlight meaningful relationships:

\begin{itemize}
	\item An increase in the volume of WMHs is associated to decreased SV and EDV, together with a smaller reduction of ESV, leading to a decrease of EF which is related to reduced contractility.
	\item Similar dynamics are associated to brain volume loss.
    \item The number of WMHs exhibits different dynamics than WMHs volume, associated to the increase in afterload and the increase of EDV.
\end{itemize}

From the plots showing the estimation of the parameters we can observe the ones driving the observed dynamics. For example, WMHs and brain volumes changes are mainly driven by the joint evolution of peripheral resistance ($R_{p}$), contractility ($\sigma_{0}$) and size of the left ventricle $(R_{0})$. The changes in the number of WMHs are related to heart-remodelling changes, driven by $R_{0}$ and by a decrease of $\sigma_{0}$ and the ventricular stiffness $c_{1}$. In supp. Figure 2 we provide analysis and discussion of the remaining conditional features (age, BSA and brain ventricles volume). Overall, in the simulated dynamics we can identify several physiological responses in line with the clinical literature. Moreover, our model may be useful to give insights in currently controversial topics such as concerning the pathogenesis of WMHs. Our results suggest that the evolution of cardiac function with respect to brain and WMHs volumes is similar to the one due to aging (see supp. Figure 2), while the effect induced by the number of WMHs is similar to the one related to ventricles enlargement. In both cases the changes in $R_{0}$, $\sigma_{0}$ and $C_{1}$ suggest that the increase in the number of WMHs and the enlargement of the brain ventricles are related to heart-remodelling processes, such as loss of contractility or decrease in compliance. These findings are in line with clinical observations \cite{Jefferson2009} relating lower cardiac output with higher burden of WMHs and reduced brain volume.

\section{Conclusion}

We presented a generative model that enables the analysis of complex physiological relationships between heart and brain in datasets where we have minimal available features. The framework allow us to emulate a lumped cardiovascular model through data-driven inference of mechanistic parameters, and provides us a generative model to explore hypothetical scenarios of heart and brain relationships. In the future, the model will allow to potentially transfer the knowledge learnt in UK Biobank to datasets where we have minimal cardiac information, to explore the relationship between brain conditions and cardiovascular factors in specific clinical contexts, such as in neurodegeneration. Our approach could also be extended to account for deep learning architectures, and the framework could be further improved by jointly accounting for multiple outputs, which are currently modelled independently, or by including spatial information from imaging data, beyond the modelling of scalar volumetric features. Furthermore, while the cardiovascular features considered in this study are rather general, more complex features will allow to study more realistic cardiovascular models. For example, while the mechanistic model used in this study does not simulate cerebral blood flow,  previous studies suggested that WMHs may be due to local vascular impairment \cite{Muller2014}. Hence, by selecting appropriate clinical features we could constrain the imputation by means of any biophysical model representing the desired aspect of systems biology. This could represent an innovative tool in real world scenarios, for which multi-modal patient data is often limited or not available.

\section*{Acknowledgments}

This work has been supported by the French government, through the 3IA Côte d’Azur Investments in the Future project managed by the National Research Agency (ANR) with the reference number ANR-19-P3IA-0002, and by the ANR JCJC project Fed-BioMed 19-CE45-0006-01. The project was also supported by the Inria Sophia Antipolis - Méditerranée, “NEF" computation cluster and by the Spanish Ministry of Science, Innovation and Universities under the Retos I+D Programme (RTI2018-101193-B-I00) and the Maria de Maeztu Units of Excellence Programme (MDM-2015-0502). This research has been conducted using the UK Biobank Resource undder Application Number 20576 (PI Nicholas Ayache). Additional information can be found at: https://www.ukbiobank.ac.uk .


%
%
\bibliographystyle{splncs04}
\bibliography{bibliography} 

\end{document}